\documentclass[twoside,leqno,twocolumn]{article}
\usepackage{ltexpprt}
\usepackage{amsmath}

\begin{document}

\title{\Large Document Structure Measure for Hypernym discovery}
\author{Aswin Kannan, Shanmukha C Guttula, Balaji Ganesan, Hima P Karanam, Arun Kumar \\
IBM Research \\
Email: (aswkanna, shguttul, bganesa1, hkaranam, kkarun)@in.ibm.com 
}
\date{}

\maketitle




\fancyfoot[R]{\scriptsize{Copyright \textcopyright\ 2018\\
Copyright for this paper is retained by authors' organization}}



\begin{abstract} \small\baselineskip=9pt Hypernym discovery is the problem of finding terms that have \emph{is-a} relationship with a given term. We introduce a new context type, and a relatedness measure to differentiate hypernyms from other types of semantic relationships. Our Document Structure measure is based on hierarchical position of terms in a document, and their presence or otherwise in \emph{definition text}. This measure quantifies the document structure using multiple attributes, and classes of weighted distance functions.\end{abstract}

\section{Introduction}
Annotating data and deciphering relationships is a key task in the field of text analytics. Hypernym discovery falls into one major embodiment of such relationship classes. More specifically, this deduces ``is-a'' relationship between any two terms in a document corpus, thereby falling into the scope of several other applications in Natural Language Processing including Entailment and Co-reference Resolution. 

A number of supervised methods have been proposed for this problem~\cite{snow-nips-2004, kamel-17}
and their efficient performance has been clearly observed. However, one of the major drawbacks is the requirement of training data in specific form and a larger volume as with all their learning subsidiaries. Say, in case of the music domain~\cite{anke2016}, document corpuses pertaining to training data change frequently, thereby rendering any supervision impractical. Similar examples can be found in regulatory compliance, where guidelines and circulars get frequently updated by bureaucrats. 

Specific to these settings, some unsupervised methods~\cite{shwartz2016hypernyms} have seen to demonstrate promising performance in select domains. One recent work~\cite{baisa15} in the context of land surveying addressed these issues in a purely rule based framework. This rests on using word similarities and co-occurrence frequency in deciphering the hierarchical relationship between any two terms. \cite{shwartz2016hypernyms} surveyed a number of relatedness measures on multiple datasets and proposed ways to choose the measures based on the dataset. More specifically, different measures such as similarity, informativeness, inclusion, and reverse inclusion were proposed and a combination of multiple such measures demonstrated better performance. \cite{chang2017distributional}
also proposed similar distributional inclusion vectors (combining multiple measures) and observed similar results.


Our framework follows the lines of~\cite{shwartz2016hypernyms}, but however focuses majorly on modeling document structure in the form of mathematical measures. In recent work\cite{yamada2009hypernym}, there has been an exploration of using hierarchical Structures for hypernym discovery, but they assume presence of Wikipedia entries for many of the candidate terms. Our document structure measures on the other hand are very general in that they encompass (but not restricted to) section titles, bulleted lists, and highlighted text. We also extend this concept of document structure to classify personal data (binary). Some examples correspond to metadata headers in JSON files. 

Our contributions are three fold: firstly, we mathematically formulate indicator based context vectors to quantify our structural measures. Secondly, based on the context vectors, we develop mathematical functions to relate word-pairs and classify personal data. Finally, it is not straightforward to manually decipher and annotate documents with respect to their document structure. We render this automation feasible by deploying System-T based rules and methods. Some primary works in the line of these rule based information extraction schemes are presented in~\cite{chiticariu2010systemt,wang2017towards}.  
Recent work~\cite{madaan2017visual} has specifically dealt with extracting titles, section and subsection headings, paragraphs, and sentences from large documents. Additionally~\cite{agarwal2017cognitive}
extracts structure and concepts from html documents in compliance specific applications. We deploy bases from~\cite{agarwal2017cognitive,madaan2017visual} to automate our process of document structure discovery and annotations.  

The rest of the paper is organized into four sections. Section~\ref{sec:back} expands on the basics of hypernym discovery and inclusion measures. Section~\ref{sec:measures} discusses in detail our document structure measures. In section~\ref{sec:implement}, we test the performance of our algorithms on the older Wikipedia data sets and compare them with results from literature. In process, we also introduce newer data sets (ENRON Email corpus) and observe the performance of our algorithms.  Initial numerical results are promising and seem to pave way for further analysis of such context vectors based on deep learning and machine learning based approaches.
We conclude in section~\ref{sec:conclusion}.

\section{Background}
\label{sec:back}
For the purpose of completeness, we restate the measures relevant to our work from past literature~\cite{shwartz2016hypernyms}. To quantify the closeness of any two words, x and y, we restate the definition of inclusion measure as follows. 
\begin{equation}
Invcl(x,y) = \sqrt{CDE(x,y)(1-CDE(x,y))}
\label{eq:reclarke}
\end{equation}
The above is extremely general in that the square root can be replaced by suitable convex functions. These can be nonlinear or even nonsmooth. 
The significance of the function is that this mimics general distance functions like the euclidean norm. CDE refers to the Clarke DE measure which is also restated as follows.
$$CDE(x,y) = \frac{\sum_{c \in C(x) \cap C(y)} \min(v_c(x),v_c(y))}{\sum_{c \in C(x)}v_c(x)},$$
where $x$ and $y$ respectively denote the words in the document and C(.) refers to the context in which these words occur. Here, we note that CDE is usually similar to a probability distribution, with an upper limit on the value to be 1.  In short, the numerator denotes the number of times words $x$ and $y$ appear together in a context $c$ and the denominator merely denotes the total number of instances in which at-least one of $x$ and $y$ appear independently. The context $c$ is extremely crucial and completely helps in ascertaining whether the words $x$ and $y$ appear together because of hypernymy relationship (or because of coincidence). 

Getting into finer details, $v_c(x)$ quantifies 
the context vector of word $x$ (also termed as embeddings). This context can refer to the window of operation, features representing Parts Of Speech (POS), character embeddings, and so on. Say, if the window of operation, $z_c^1(x) = 4$, then all words that appear within a distance of 4 to the left and right of the word $x$. This can obviously refer to words of different dimensions and this can also be generalized to individual characters also depending upon the requirement. Even generalizations exist where this can be kept as a sliding window, the size of which can change from one word to another (depending completely on the domain and sub-domain). In cases where repetitions occur (in all possible contexts not necessarily restricted to the above), averaging amongst samples is encouraged. Alternatively, logarithmic probability functions can also be deployed. 


From the above concepts of word embeddings, we derive quantifiers for document structure to predict hypernym relationships. While this is discussed in detail in the next section, we set a prelude in terms of the mathematics behind coming up with those ideas. For the present section for the purpose of notational clarity, we do not disturb the specification of the vector $v$. We instead add a new vector $z$ to specify additional dimensions. 

To begin with, let us consider the instance of section titles. Motivated by the fact that hypernyms can be found in the section titles with a higher probability than hyponyms, we specify functionals that capture such cases. As a simple example, the word ``Country'' or ``Geographical Location'' is a good qualifier for a header, whereas South-East Asia and ``Northern America'' are less probable to be headers. On a side node, if South-East Asia forms the part of a section title, it is more likely for the document title to contain the word ``Country''.
This can be expressed in a mathematical sense, where a word appearing in a section title proportionately increases the probability of it being a hypernym. The inverse relationship is defined by an appropriate fraction. Let $z_c^2(x)$ denote the context specific to section titles for a word $x$. Then, the distribution function can be expressed as follows.
\begin{equation}
CDE_c^2(x,y) = w_3 z_c^2(y) + \frac{w_4}{z_c^2(x)}.
\label{eq:secdimension}
\end{equation}
Description of text contain significant number of hypernyms. Say, the word color can be describe blue or green and used repetitively in text to establish certain relationships or define 
key points in a discussion. Say, in historical text, political party names can be used quite often to describe a scenario. These obviously qualify as hypernyms with the composite elements like leaders, affiliates, and most importantly the sub-classes in multiple nations. Say, the democratic party may be used to describe some text and can refer to different names in different countries. However, we note one caveat and observe that even hyponyms can be found in description text. Let $z_c^3(x)$ denote the context specific to definition text. To clearly state the higher probability of the former, we come up with the following logarithmic function.
\begin{equation}
CDE_c^3(x,y) = w_5 z_c^3(y) \log{z_c^3(x)}.
\label{eq:thirddimension}
\end{equation}
Note that all the weights $w_1, w_2, w_3, w_4$, and $w_5 \geq 0$ are nonnegative. At this point, we clearly note that the expressions used above are very rudimentary and merely help in clear understanding of how hypernymy measures should be constructed based on document structure. For a deeper dive and strong theoretical and numerical conceptualization, we move to the forthcoming section.  

\section{Deploying Structure}
\label{sec:measures}
Prior to defining our mathematical model, it is important to observe that some words $x$ can be hypernyms in general, without being specifically associated to a word. These words form the top portions of hierarchies. Say, for instance ``personal information'' can point to names, addresses, biographical details etc. and covers a very broad range of labels. While it is true that these cannot be called hypernyms without the presence of the required subsets, we apportion a value of probability independently to such generalized terms. Next, we define vectors, $v_a(x)$, $v_b(x)$, and $v_c(x)$ to define the contexts in which a word $x$  can be a relational hypernym, relational hyponym, and general hypernym respectively. The contexts directly correspond to document structure and we additionally note that $v_a$ and $v_b$ have the same contexts, but in different (opposite) senses.  
\subsection{Relational Context Vectors:}
We start with an example of bulleted text. As we have context windows, here we split the entire document into multiple paragraphs ensuring that each paragraph at the most contains only one set of bulleted text. It is obvious to note that bulleted text are more probable to be hyponyms�  (Lists / Enumerations also included). Consider the instance
of text below. 

\textit{``X contains the following:}
\begin{itemize} 
\item \textit{X1}
\item \textit{X2''} 
\end{itemize}
Here, X1 and X2 are hyponyms of X.
Given two words $x$ and $y$, their probability of hypernym-hyponym relationship specific to a context $i$ (in this case bulleted list) can be stated as follows:
$$\rho_i(x,y) = \frac{\sum_{j=1}^{m_i}v_a^{i,j}(x) \cap v_b^{i,j}(y)}{\sum_{j=1}^{m_i}v_a^{i,j}(x)}.$$
In the above expression $j$ refers to an entity mention, which in this case is the presence of the word $x$ in paragraph $j$ that contains a bulleted list. In case a paragraph $j$ does not contain a bulleted list, the corresponding entries are 0 for both the numerator and denominator sub-portions. In some cases a bulleted list may contain more than one occurrence of a hypernym / word. We merely consider the above expression to have an indicator function and do not pursue on the track of multiple occurrences.  The vectors $v_a$ and $v_b$ are specified in an opposite sense. If a word occurs at the text preceeding the bullets, they are directed towards the indicator function in $v_a$ and if they occur within the sub-bullets, those are attributed towards the indicator function in $v_b$. Usually in such cases if $v_a^{i,j}(x) = 1$, then $v_b^{i,j} = 0$. However, it can also be true that some words can be present in both the preceeding text or sub-bullets, leading to both $v_a$ and $v_b$ taking the value of 1. 

Generalizing the above to all possible contexts, we have the following.
$$\rho(x,y) = \sum_{i=1}^{n} w_i f_i(n_x,n_{i,x}) \rho_i(x,y),$$
where $w_i$ refers to the weight assigned to each context (pre-set by the user depending on the application) and $f_i(.)$ denotes an importance function corresponding to the occurrence of the entities both in the presence of the context and overall (presence and absence included). More specifically, $n_{i,x}$ refers to the number of times the word $x$ has occurred in text preceeding the bullets in the document and $n_x$ denotes the overall number of times the word $x$ has appeared in the document. 
Besides the bulleted list, the other contexts that we consider in this work are the following:
\begin{itemize}
\item Hyperlinks / URL content is more probable to contain hypernyms in the first portion and hyponyms in the second portion.

\textit{Eg: www.webmd.com/..../symptoms/headache}. Note that symptoms goes into the $v_a$ bin and headache into $v_b$. 
\item Footnotes are more probable to contain hyponyms. 

Eg: Let us consider $Word^{1}$. Here, ``Word'' is the Hypernym and the footnote corresponding to ``$1$'' can contain hyponyms. 
\item Section Headers / Paragraph headers / Subsections follow hierarchical order. Say, when a word x occurs in the section title and word y occurs in the paragraph, x is more probable to be a hypernym of y. 
\item Words within brackets are more probable to be hyponyms. 

Eg: Eastern Geographical Location (Say, Japan, Singapore, and Thailand).
\item Subscripts and Superscripts are more probable to be hyponyms. 

Eg: $1^{st}$, $2^{nd}$, and $3^{rd}$ Class. Here, class is the hypernym.
\item Words succeeding Indents are very probable to be hypernyms (very similar to section titles). First few words after an indent denote an opening and can contain a hypernym.

Eg: \hspace{2mm} A few \textbf{priorities} are required. Say, evening exercises, Yoga, and jogging help maintaining fitness. 
\item Words defining Under-braces and Over-braces are usually Hypernyms (Say in mathematical descriptors)

Eg: {The following} $\underbrace{Expression_1, Expression_2, Expression_3}_{Hypernym}$ {are quite helpful in figuring out the essence of this article.}
\end{itemize}

\subsection{General Context Vectors} The expression of the context vector differs slightly in this generalized case. 
While we look at pairs, these are only specific to occurrences of the hyponym within a window of $x$. In this case,
the vector $v_c$ covers and accounts for both the hypernym and hyponym. Say, when a word's context indicates a hypernym, 
the corresponding value of $v_c$ is set to 1. In cases of hyponyms, the value is set to -1 and 0 otherwise. The probability measure 
takes the following form: 
$$ \rho_i(x,y) =  \sum_j v_c^{i,j}(x) \left(\frac{1-\min(v_c^{i,j}(y),0)}{1+\max(v_c^{i,j}(y),0)}\right).$$
The expression including the weights defining the complete function $\rho(x,y)$ follows the same pursuit as earlier.
In this work we consider the following contexts for general vectors.
\begin{itemize}
\item Captions of figures and tables are more probable to contain hypernyms.
\item Text with hyphens, semicolon, commas, quotations are more probable to contain hypernyms.  
\item Word preceeding a question mark if a noun is more probable to be a hyponym.
\item Words within Markings / Watermark / Highlighting are more probable to be hyponyms / confidential information.
\item Single worded cells in excel like data are more probable to contain hypernyms. 
\item When looking at shapes, the first few boxes would correspond to hypernyms. As an example, this is quite common in MS-Visio. 
\item Upper cased words are more probable to be hypernyms. 
\item Color / Bold / Italics / Underline are very similar to highlighting and are usually hypernyms. 
\item Words after symbols ``$>, <, ||, \& \&, \textrm{and} \#$'' are usually hyponyms. 
\item Words corresponding to Info-Boxes / Remarks are usually hypernyms (examples from Wikipedia).
\item Higher Font-Sized text are usually hypernyms. 
\end{itemize}
Some additional contexts (with considerably lesser weights) in a minor sense are also considered and described as follows:
\begin{itemize}
\item More Number of words in a cell in an excel file indicates higher probability of such words to be hyponyms. 
\item Words in Introduction / Conclusion are likely to be hypernyms.
\item References are more probable to contain hyponyms.
\item Appendix based text contain more hypernyms. 
\item Double Spaced text contain more hypernyms generally (eg. double spacing for quotations ? reported conversations).
\item Keywords / Abstract / General terms generally constitute hypernyms (Very common in journal articles). 
\end{itemize}

\subsection{Personal Data Extraction}
While the scope of this paper is restricted to hypernymy, this research has a great potential 
value in general relationship discovery, say as examples personal data tagging and meronyms. 
Several data protection regulations demand extracting personal data entities from large document corpuses. Metadata in the notion of unstructured text is very helpful in finding out whether some portions contain sensitive data (say biometrics, genetic information, and credit card numbers). In this regard, for the purpose of completeness, we analyze document structure properties and state the following contexts to define measures in a very similar flavor as earlier.
\begin{itemize}
\item Special characters such as ``****, xxx'' are usually associated with sensitive information.
\item Attachments with special names or numbers can include sensitive data.
\item As opposed to plain text, the probability of finding sensitive data in tables in much higher. 
\item Section titles are very helpful in finding information about the context and in turn the possibility of personal data.
\item Footers of emails can contain personal data.
\item Responses to questionnaires and texts within blanks can point to sensitive data.
\item Boxed or colored (Say red) text can contain sensitive data.
\item Indented text and larger sized textual portions can contain sensitive data (say codes / PNR).
\end{itemize}
The corresponding measures can be constructed as earlier in the case of hypernymy.

\section{Implementation}
\label{sec:implement}
The intent of our numerical results is three fold. First, we note that the base code from~\cite{shwartz2016hypernyms} predicts hypernymy relationship between any two terms, whereas our work predicts hypernyms for any specific input term $x$. Our first numerical contribution is from the standpoint of SemEval, where the task is to predict such hypernyms (more details below).  Note that we included bigrams as opposed to unigrams and used Spacy to generate POS tags as required by the original implementation. We further mention that these are very general structure theoretic measures that helps in hypernymy detection. 
Second, we try to show the enhancement in hypernymy detection by using our proposed measures. Second, we introduce a new data set (ENRON) as a path for future study and run a portion of our exercises on the same. All the computation was done on a Linux Cluster with 8 cores and 4 GB RAM. We explicitly state that our implementation is built on top of publicly available code from ~\cite{shwartz2016hypernyms}. As mentioned by the authors of the corresponding base code, no one measure performs best on all datasets.   Our document structure measures are built as wrappers (in Matlab and Python) around their base.
\subsection{SemEval}
In the SemEval tasks, since we are provided with vocabulary words file, which contains the exhaustive list of all possible hypernyms, we used them for prediction. We filtered the vocabulary words file to remove few corrupted words and very small words. We restricted our vocabulary to words provided to us but we did not use a minimum frequency filter as done by the original authors. In the original implementation of \cite{shwartz2016hypernyms} two words are compared at time. We vectorized this scoring task by comparing each word at a time with all the possible hypernyms. Also, we used the different data structures for more efficiency and reduce latency. Once we got the scores for a given word, we ranked the words and produced top 100 words as hypernyms.

\paragraph{Data Pre-processing}
We worked on the following two sub-tasks in the SemEval 2018 Hypernym Discovery Task 9: 
\begin{itemize}
    \item Music
    \item Medical
\end{itemize}

The corpus data provided by SemEval for each sub-task is unstructured data. It is in the form of untagged sentences with one sentence per line and provided to us in the form of text file. We used Spacy[citation needed] to parse each sentence of corpus and tagged each word with it's lemma and POS tag. We then used the tagged corpus for training. We considered only the Nouns, Verbs and Adjectives by filtering out the remaining word types. We also restricted our corpus vocabulary to the words in the Vocabulary file provided to us.

\paragraph{Distributional Space}
In \cite{shwartz2016hypernyms}, two parameters are described, Context Type and Feature Weighting. We experimented with different combinations of Context Type and Feature Weighting. We found the window-based context and PMLI feature weighting to work better on the Music and Medical datasets. Our evaluation on both the datasets are presented in Tables 1 and 2. 

\begin{table}[]
\centering
\caption{Music Dataset}
\label{music-table}
\begin{tabular}{|l|l|l|l|l|}
\hline
\textbf{\begin{tabular}[c]{@{}l@{}}Context\\ Type\end{tabular}} & \textbf{Measure} & \textbf{MAP} & \textbf{MRR} & \textbf{P@5} \\ \hline
win5                                                            & ClarkeDE         & 0.205        & 0.089        & 0.093        \\ \hline
win5                                                            & invCL            & 0.197        & 0.088        & 0.092        \\ \hline
win5d                                                           & ClarkeDE         & 0.217        & 0.094        & 0.097        \\ \hline
win5d                                                           & invCL            & 0.211        & 0.093        & 0.096        \\ \hline
\end{tabular}
\end{table}

\begin{table}[]
\centering
\caption{Medical Dataset}
\label{medical-table}
\begin{tabular}{|l|l|l|l|l|}
\hline
\textbf{\begin{tabular}[c]{@{}l@{}}Context\\ Type\end{tabular}} & \textbf{Measure} & \textbf{MAP} & \textbf{MRR} & \textbf{P@5} \\ \hline
win5                                                            & ClarkeDE         & 0.204        & 0.091        & 0.091        \\ \hline
win5                                                            & invCL            & 0.207        & 0.093        & 0.09         \\ \hline
win5d                                                           & ClarkeDE         & 0.22         & 0.097        & 0.099        \\ \hline
win5d                                                           & invCL            & 0.222        & 0.1          & 0.103        \\ \hline
\end{tabular}
\end{table}

\subsection{Document Structure Measures:}
We note that document structure measures can also be referred to as dependency based contexts and can be suitably deployed with dependency parse trees. 
We consider a set of four document structure features, namely section titles, footnotes, subscripts and superscripts, and captions for our study. The measures are tested on both datasets, namely Wackypedia (as the earlier subsection) and ENRON (emails). We test measures one at a time and all at once for Hypernyms and report the results for the two datasets as follows.
\paragraph{Wikipedia Corpus:}
Here, we used an unsupervised scoring measure to determine whether a word $y$ is a hypernym of $x$. We tried different measures mentioned in \cite{shwartz2016hypernyms}. We observed that invCL and ClarkeDE are the best performers for Medical and Music Datasets respectively. Distributional inclusion hypothesis states that the prominent contexts of a hyponym($x$) are expected to be included in its hypernym($y$). Using the best inclusion measures we found, given a word/hyponym, we score all vocabulary words for possible hypernyms and rank them. We take the top $k$ scorers and output them as hypernyms $H_w$ for a given word $w$. We are proposing a new measure for Hypernym Discovery based on the heuristic that expanding the context window with additional relevant words will improve the vector representation in a way to better distinguish hypernym relations from other relations. In the rest of the section, we describe different Document Structure based contexts, and provide the mathematical definition of the measure.
\paragraph{ENRON Emails:}
We observe that hypernyms tend to more general, while hyponyms are more specific in defining a real word entity. We propose that the section and document headings tend to generalize the description of real world entities. PMLI as described in 3.2 gives the conditional probability of a term being a hypernym, given its co-occurence with another term. We observe that this probability increases when one of the terms is a generalized term that typically appears in document and section headings.
Drawing from the field of information retrieval, we observe that text that describes a term, called definition text, is more likely to contain hypernym terms than any other text in the document. There is related work on identifying if a text is definition text. It might also be possible to assume that the first paragraph in a Wikipedia page, is more likely to describe a term, and hence can be considered a definition text. 
We plan to leverage work in the areas of text summarization to reduce the noise in our Document Structure measure.

\section{Conclusion}
\label{sec:conclusion}

We observe that the findings of \cite{shwartz2016hypernyms} hold true on domain specific datasets that we experimented with. We found that there is not a single measure which impacts Hypernym Discovery. 
We have introduced a new relatedness measure, based on Document Structure to distinguish Hypernym relations from other kinds of semantic relations between two terms. Besides incremental performance, we see some good new predictions of hypernyms that were otherwise absent using standard contexts and measures in literature. Our analysis on relatively newer datasets like ENRON are in sync with real-world applications and helps with directions of future research.

\end{document}